\documentclass[10pt, a4paper]{article}

\usepackage{lrec2026} 
\usepackage{dirtytalk}
\usepackage{hyperref}
\usepackage{amsmath}
\usepackage{amsfonts}
\usepackage{booktabs}

\usepackage{cjhebrew} 

\usepackage{xcolor}

\newcommand{\diacness}[0]{diacritifulness}

\title{The Degree of Language Diacriticity and Its Effect on Tasks}

\name{Adi Cohen, Yuval Pinter} 

\address{Institute for Applied AI Research \\
         Ben-Gurion University of the Negev \\
         Beer Sheva, Israel \\
         \texttt{\{adibc@post, uvp@cs\}.bgu.ac.il}}

\abstract{
Diacritics are orthographic marks that clarify pronunciation, distinguish similar words, or alter meaning.
They play a central role in many writing systems, yet their impact on language technology has not been systematically quantified across scripts.
While prior work has examined diacritics in individual languages, there's no cross-linguistic, data-driven framework for measuring the degree to which writing systems rely on them and how this affects downstream tasks.
We propose a data-driven framework for quantifying diacritic complexity using corpus-level, information-theoretic metrics that capture the frequency, ambiguity, and structural diversity of character-diacritic combinations.
We compute these metrics over 24 corpora in 15 languages, spanning both single- and multi-diacritic scripts.
We then examine how diacritic complexity correlates with performance on the task of diacritics restoration, evaluating BERT- and RNN-based models.
We find that across languages, higher diacritic complexity is strongly associated with lower restoration accuracy.
In single-diacritic scripts, where character-diacritic combinations are more predictable, frequency-based and structural measures largely align.
In multi-diacritic scripts, however, structural complexity exhibits the strongest association with performance, surpassing frequency-based measures.
These findings show that measurable properties of diacritic usage influence the performance of diacritic restoration models, demonstrating that orthographic complexity is not only descriptive but functionally relevant for modeling.
 \\ \newline  }

\begin{document}

\maketitleabstract

\section{Introduction}

Diacritics have been known to affect performance of language technology systems, but a thorough data-driven survey has yet to be done.
\citet{gorman-pinter-2025-dont} started to explore this issue, presenting anecdotal evidence and providing rules of thumb for NLP practitioners.

In this study, we consider the presence of diacritics a property of the writing system of a language and build a bottom-up framework for quantifying the degree to which diacritics play a role in corpora of a language, leading to a mechanism for examining the \emph{degree} to which diacritics affect task performance, for a task expected to present such variation, namely diacritization itself.

Our approach considers \say{\diacness{}} of a langauge along a spectrum:
for example, in Spanish vowels occasionally feature one kind of an accent mark (e.g.,~\texttt{\'a}), and one base consonant character (\texttt{n}) is shared by two phonemes (\texttt{n} and \texttt{\~{n}}).
In Vietnamese, both vowel quality and tone are marked with co-inciding marks.
In Hebrew, vowels are marked on phonetically preceding consonant characters and exist in large combinatorial distribution.
Once this variation is accounted for quantitatively, the true effects of a language writing system on technology performance can be examined via tools like correlation.

Concretely, we propose the first-ever attempt to quantify the degree to which a language uses diacritics in its canonical writing system.
Our approach is data-driven and information-theoretic: we calculate corpus-level metrics based on the distributions of diacritic marks, characters, and their combinations in each language.
We assess the effect this property of \say{\diacness{}} has on the closest task to the subject matter: \emph{diacritization} of undiacritized text, measured by word-level and character-level accuracy.
Our results, identifying the correlation between a language's \diacness{} and success of various neural diacritization models, suggest that consideration of whether each of a script's characters admits one or more diacritic at a time, which we term single- vs. multi-diacritic systems, affects this relationship greatly, particularly for transformer-based diacritization models.
We hope that our framework encourages more work on gauging written properties of languages and their effects on tasks, perhaps more downstream ones such as question answering and machine translation.\footnote{Our code, models and data are available at \\ \url{https://github.com/MeLeLBGU/Diacriticity}.}

\section{Background}

Diacritics are marks attached to base letters that encode additional phonological and/or grammatical information. They may indicate vowel quality or length, tone, gemination, or morphological distinctions. In some languages, like Vietnamese, diacritics are essential and form part of standard spelling. In other languages, such as Arabic and Hebrew, diacritics encode important phonological or grammatical distinctions but are often omitted in everyday writing. Writing systems that omit some or all diacritic marks are often referred to as defective writing systems, since the written text does not fully specify pronunciation or lexical identity.

The omission of diacritics introduces ambiguity. A single undiacritized form may correspond to multiple valid interpretations, requiring the inference of the intended meaning from context. This has been cited as one major reason that speech and language technologies for Arabic and Hebrew seemingly lag behind other, similarly-resourced languages~\citep[e.g.,][]{tsarfaty-etal-2019-whats}.

In text-based systems, including machine translation, different diacritizations may have several valid lexical or grammatical meanings. Despite this, many NLP pipelines remove diacritics during preprocessing, especially in multilingual transformer-based models such as mBERT (\citealp{clark-etal-2022-canine}).

Diacritization, the task of restoring missing diacritics, has been extensively studied, particularly for Arabic and Hebrew. Early approaches relied on rules, lexicons or morphological analyzers 
(e.g.,~\citealp{darwish-etal-2017-arabic,krstev-etal-2018-knowledge}). 
More recent work formulates diacritization as a sequence tagging or sequence-to-sequence problem using neural architectures such as RNNs and transformers 
(e.g.,~\citealp{belinkov-glass-2015-arabic,shmidman-etal-2020-nakdan,gershuni-pinter-2022-restoring,naplava-etal-2018-diacritics,naplava-straka-strakova-2021}).

While these models perform well within individual languages, most work remains focused on language-specific or single writing system, such as Latin-based scripts.
As a result, cross-script comparisons remain limited.
There has also been relatively little systematic investigation of how orthographic properties themselves, such as diacritic frequency, structural diversity or ambiguity, affect restoration difficulty across languages.

Languages differ in how diacritics are used.
In some scripts, each character can take only a single diacritic, while in others, base characters can carry multiple simultaneous marks.
We refer to characters carrying exactly one diacritic as \textbf{single-diacritic}, and to those carrying two or more simultaneous marks as \textbf{multi-diacritic}.
At the language level, we classify a writing system as multi-diacritic if it permits any character to bear multiple simultaneous diacritics; otherwise, we classify it as single-diacritic.

Beyond frequency and function, languages also vary in the predictability and diversity of their diacritic patterns.
These differences suggest that diacritic usage varies along a spectrum of structural complexity.
We propose quantitative corpus-level metrics to capture this variation and examine how it relates to neural diacritization performance across languages.

\section{Metrics}
\label{sec:metrics}


\begin{table*}
    \centering
    \begin{tabular}{lrrrrrr}
    \toprule
        \textbf{\say{Corpus}} & \textbf{Chars} & \textbf{Diacs} & \textbf{Uniq. Diacs} & \textbf{RS} & \textbf{DTS} & \textbf{DSS} \\
        \midrule
        El niño bebió café en la mañana & 25 & 4 & 2 & 0.28 & 0.15 & 0.11\\
        \normalfont \cjRL{hay*ElEd +sAtAh qApEh b*ab*oqEr} & 14 & 14 & 6 & 0.33 & 0.54 & 0.52\\
        \bottomrule
    \end{tabular}
    \caption{
    \label{tab:examples} Example sentences in Spanish and Hebrew illustrating how diacritic-based metrics are computed. The table reports the number of characters, total diacritics, number of unique diacritics, and the resulting RS, DTS, and DSS values.}
\end{table*}

Let $C$ be the set of base characters of a language (e.g., the Latin letters \texttt{a}--\texttt{z}), and let $D$ be the set of diacritic marks that may attach to them.
We define a \emph{rune} as a base character together with zero or more diacritics, so that both unmarked and diacritized characters are treated uniformly.
For example, the Spanish character \texttt{{\'a}} can be viewed as the base letter \texttt{a} with an acute accent, while the Hebrew form \<b*A> consists of the base letter \<b> with two diacritic marks \<\dottedcircle *> and \<\dottedcircle A>.
In both cases, we treat the base character together with its diacritics as a single rune.
A diacritized string is then a sequence of runes $r_1, \ldots, r_n$, where each rune $r_i$ consists of a base character $c_i \in C$ and a (possibly empty) subset of diacritics from $D$.

We follow previous work in defining a stripping function $\sigma$ that removes diacritics and returns the corresponding base character sequence $c_1, \ldots, c_n$.


To capture the deeper properties of writing scripts and diacritics, we introduce a set of information-theoretic metrics that quantify the relationship between base characters and their diacritized realizations.

We present examples for our metrics on Hebrew and Spanish in \autoref{tab:examples}.

\paragraph{Rune Surprisal.}

Rune surprisal (RS) quantifies the level of uncertainty associated with a diacritized version based on its base character.
It is high when a character appears with multiple competing diacritic forms of similar probability, and low when a single form clearly dominates.

Formally, let $r$ be a rune with base character $\sigma([r])=c$, we define:
\[
P(r \mid c) = \frac{\#(r)}{\sum_{r' : \mathrm{base}(r') = c} \#(r')}.
\]

The suprisal of a rune is:
\[
RS(r) = -\log P(r \mid c).
\]

Languages that show higher average rune surprisal when calculated over a corpus display greater ambiguity per character, which we expect to make diacritization more difficult.

\paragraph{Diacritic Token Surprisal.}

Diacritic token surprisal (DTS) breaks each rune into its individual diacritics marks and measures how unexpected those marks are given the base character.
Rather than treating the full diacritized form as a single unit, this metric evaluates the contribution of each individual mark.

Let $d$ be a diacritic mark appearing as part of a rune $r$ with base character $c$, We define:
\[
P(d \mid c) = \frac{\#(d,c)}{\sum_{d'\in D} \#(d',c)},
\]
where $\#(d,c)$ is the frequency of mark $d$ with character $c$.

The diacritic token surprisal of rune $r$ is:
\[
DTS(r) = - \sum_{d \in r} \log P(d \mid c).
\]

This metric captures token-level diacritic unpredictability. It is especially informative in systems where multiple diacritics can occur on the same character at the same time.

\begin{table*}[t]
\centering
\small
\setlength{\tabcolsep}{3pt}
\begin{tabular}{lrrrrrrl}
\toprule
\textbf{Language} &
\textbf{Diacritic} &
\textbf{Multi} &
\textbf{\% Words} &
\textbf{\% Lines} &
\textbf{Mean} &
\textbf{\# Runes} &
\textbf{System} \\
 \textbf{\phantom{Ger} Source} & \textbf{Density \%} & \textbf{Diacritics \%} & & &
 \textbf{Diacs/Word}
 \\
\midrule
\textbf{German} & 1.357  & 0.000  & 8.011  & 67.021  & 1.017  & 3   & Single \\
\multicolumn{5}{l}{\phantom{Ger} UD 2.15 (GSD), 2.10 (HDR)} \\
\textbf{Spanish}             & 2.336  & 0.000  & 11.349 & 86.454  & 1.008  & 7   & Single \\
\multicolumn{5}{l}{\phantom{Ger} UD2.10 (AnCora), 2.9 (GSD)} \\
\textbf{Croatian}            & 2.607  & 0.000  & 13.812 & 73.601  & 1.052  & 4   & Single \\
\multicolumn{5}{l}{\phantom{Ger} \citet{naplava-etal-2018-diacritics}} \\
\textbf{Galician}            & 2.984  & 0.000  & 15.894 & 71.931  & 1.003  & 6   & Single \\
\multicolumn{5}{l}{\phantom{Ger} UD 2.12 (CTG)} \\
\textbf{Portuguese}          & 3.700  & 0.000  & 16.229 & 64.003  & 1.140  & 12  & Single \\
\multicolumn{5}{l}{\phantom{Ger} UD 2.13 (GSD, Cintil)} \\
\textbf{French}              & 3.845  & 0.000  & 17.539 & 90.684  & 1.119  & 13  & Single \\ 
\multicolumn{5}{l}{\phantom{Ger} UD 2.7 (GSD), 2.2 (FTB)} \\
\textbf{Romanian}            & 5.999  & 0.000  & 29.488 & 84.026  & 1.141  & 5   & Single \\
\multicolumn{5}{l}{\phantom{Ger} UD 2.8 (RRT, SiMoNERo)} \\
\textbf{Turkish}             & 6.510  & 0.000  & 30.971 & 74.863  & 1.347  & 9   & Single \\
\multicolumn{5}{l}{\phantom{Ger} UD 2.8 (Penn, Kenet)} \\
\textbf{Lithuanian}          & 7.155  & 0.000  & 39.789 & 94.730  & 1.010  & 9   & Single \\
\multicolumn{5}{l}{\phantom{Ger} UD 2.8 (Alksnis)} \\
\textbf{Latin}               & 11.118 & 0.000  & 53.338 & 96.053  & 1.207  & 6   & Single \\
\multicolumn{7}{l}{\phantom{Ger} Chapters 1--6 of Vergil's Aeneid from Pharr's reader, digitized by Kyle Gorman} \\
\textbf{Czech}               & 14.505 & 0.000  & 55.119 & 93.514  & 1.510  & 15  & Single \\
\multicolumn{5}{l}{\phantom{Ger} UD 2.15 (CAC, PDTC)} \\
\textbf{Vietnamese}          & 25.216 & 10.200 & 83.090 & 99.571  & 1.480  & 64  & Multi \\
\multicolumn{5}{l}{\phantom{Ger} \citet{naplava-etal-2018-diacritics}} \\
\textbf{Bengali}             & 58.907 & 4.826  & 92.467 & 100.000 & 2.439  & 390 & Multi \\
\multicolumn{7}{l}{\phantom{Ger} Leipzig~\citep{goldhahn-etal-2012-building}: Bengali News 2020 100k, Wiki 2021 100k} \\
\textbf{Hebrew}              & 66.243 & 14.096 & 99.823 & 99.761  & 3.607  & 325 & Multi \\
\multicolumn{5}{l}{\phantom{Ger} Nakdimon~\citep{gershuni-pinter-2022-restoring}: Lit, news} \\
\textbf{Arabic}              & 82.813 & 9.297  & 99.552 & 89.221  & 3.686  & 353 & Multi \\
\multicolumn{5}{l}{\phantom{Ger} Tashkeela~\citep{zerrouki2017tashkeela}} \\
\bottomrule
\end{tabular}
\caption{Corpus-level diacritic usage across languages. For languages represented by multiple corpora, statistics are averaged across corpora. The columns present overall diacritic density (ordering key), proportion of multi-diacritic characters, percentage of words and lines that are diacritized, mean diacritics per \textbf{diacritized} word, corpus-wide number of distinct character--diacritic combinations (runes), and classification as single- or multi-diacritic writing systems.}
\label{tab:diacritic_stats}
\end{table*}

\paragraph{Diacritic Structural Surprisal.}

Diacritic structural surprisal (DSS) measures structural competition independently of frequency.
Instead of asking how often a mark appears, it asks how many distinct diacritized forms it appears in.

For a base character $c$, let $T(c)$ denote the set of distinct runes formed from $c$, and let $T_d(c) \subseteq T(c)$ denote the subset of those runes that contain diacritic $d$. We define:
\[
P_{\delta}(d \mid c) =
\frac{|T_d(c)|}{|T(c)|}.
\]

The structural surprise of rune $r$ is:
\[
DSS(r) = - \sum_{d \in r} \log P_{\delta}(d \mid c).
\]

This metric captures type-level structural complexity.
In single-diacritic systems, structural and token-level effects tend to align. In multi-diacritic systems, they can diverge substantially.

\paragraph{Diacritic Density.}

As a baseline surface measure, we define \emph{diacritic density} as the proportion of diacritic tokens relative to base character tokens in the corpus:
\[
\mathrm{Density} =
\frac{\sum_{d} \#(d)}
{\sum_{c} \#(c)}.
\]

This metric captures the overall diacritic load of the writing system, without conditioning on character identity or structural competition. While density does not measure ambiguity directly, it serves as a baseline indicator of how heavily diacritics are used in the writing system.

\section{Data}


We perform experiments and statistical analysis on fifteen languages across various scripts and typological characteristics.
We include Latin-derived writing systems along with scripts like Arabic, Hebrew, and Bengali.
For various languages, we provide multiple corpora.
Our dataset includes 24 corpora, sourced from a mix of Universal Dependencies~\citep[UD;][]{de-marneffe-etal-2021-universal} and extensive web corpora.
All corpora are fully diacritized.
To the best of our knowledge, all diacritics have been manually added to the texts by either original authors or professionals, with the exception of some portions of the Hebrew data which were semi-automatically diacritized using Dicta's Nakdan API~\citep{shmidman-etal-2020-nakdan} followed by manual correction (see \citet{gershuni-pinter-2022-restoring}).


To ensure comparability across languages and corpora, we implemented a fixed-size sampling approach.
For every corpus, we sample roughly 300,000 characters of diacritized text.
Sampling occurred at sentence level: corpus sentences were shuffled and gradually accumulated until the target character count was achieved.
For corpora that are smaller than the target (e.g., Latin), all available data was used and augmented by resampling sentences as needed.

Across various scripts, the mapping between characters and diacritics differs.
In certain writing systems, a rune aligns with a single Unicode code point.
For example, some Latin exist as precomposed characters: \texttt{\'a} (U+00E1, `Latin Small Letter A with Acute').
Even letters with two diacritics may be encoded as a single code point, such as \texttt{ắ} (U+1EAF, `Latin Small Letter A with Breve and Acute').
In other scripts, a rune is represented as a base character followed by one or more combining marks.
For example, the Hebrew rune \<b*A> consists of the base character U+05D1 (HEBREW LETTER BET) together with U+05BC (HEBREW POINT DAGESH OR MAPIQ) and U+05B8 (HEBREW POINT QAMATS).
Although this sequence is encoded as three unicode code points (\<b> + \<\dottedcircle *> +  \<\dottedcircle A>), it functions as a single letter with diacritics.

For scripts where diacritics are encoded as combining marks (Arabic, Hebrew, Bengali), we normalize all text to a consistent decomposed representation and compute statistics with respect to the underlying base character. Importantly, we treat each base character together with its associated diacritics as a single orthographic unit (rune), rather than as separate characters, even though they are encoded as multiple code points in Unicode.

\subsection{Diacritic Usage Statistics}
The statistics in \autoref{tab:diacritic_stats} characterize the frequency, distribution, and combinatorial diversity of diacritics across writing systems.
However, they do not directly capture predictability or ambiguity in character-diacritic mapping, which we aim to quantify using the information theoretic metrics introduced in Section~\ref{sec:metrics}.

\section{Experimental Setup}


We evaluate two established diacritization architectures in order to examine how properties of writing systems relate to model behavior across scripts.
Our goal is not to propose a new modeling approach, but to analyze performance variation as a function of script-level characteristics.

\paragraph{BERT-based Diacritizer.}

We use the BERT-based~\citep{devlin-etal-2019-bert} sequence labeling model introduced by \citet{naplava-straka-strakova-2021}, which has been shown to achieve strong performance on Latin-script languages.
Separate models were trained for each corpus in our dataset. 
For Latin-based scripts, we apply the original implementation without modification.
For non-Latin scripts (Arabic, Hebrew, Bengali), we adapted the pipeline to work with writing systems in which the diacritics are encoded as separate combining marks rather than as precomposed characters.
The transformer architecture, tokenization strategy (character-level), and training regime remain as in the original work.

For languages represented by multiple corpora, we additionally conducted cross-corpus evaluation.
The model was trained on one corpus and evaluated on a different corpus from the same language in order to check for domain robustness.

\paragraph{RNN-based Diacritizer.}

For Latin-script languages only, we evaluate the character-level RNN model of \citet{naplava-etal-2018-diacritics}, designed specifically for Latin alphabets where diacritics are precomposed in Unicode.
Its encoding assumptions do not extend to Arabic, Hebrew and Bengali, so we restrict it to Latin-script corpora.

Both models use the original training objectives and decoding strategies from the reference implementations.

\subsection{Evaluation Metrics}

We evaluate diacritization performance using two complementary accuracy measures:
\begin{itemize}
\item \textbf{Word-level accuracy}, defined as exact match of full word's diacritization.
\item \textbf{Rune-level accuracy}, defined as correct restoration of individual character--diacritic combinations.
\end{itemize}

Word-level accuracy reflects end-user correctness in downstream applications, while rune-level accuracy provides a more fine-grained view of model behavior and error patterns within words.

\section{Results}

\begin{table*}[t]
\centering
\small
\setlength{\tabcolsep}{3pt}
\begin{tabular}{llrrrrrrrr}
\toprule
\textbf{Language} &
\textbf{Family} &
\textbf{Density} &
\textbf{RS} &
\textbf{DTS} &
\textbf{DSS} &
\textbf{RNN-Word} &
\textbf{RNN-Rune} &
\textbf{BERT-Rune} &
\textbf{BERT-Word} \\
\midrule
German      & Indo-Euro  & 1.374  & 0.044 & 0.031 & 0.009 & 94.245 & 99.088 & 99.045 & 94.888 \\
Spanish     & Indo-Euro  & 2.330  & 0.092 & 0.069 & 0.018 & 91.513 & 98.426 & 98.299 & 94.038 \\
Croatian    & Indo-Euro  & 2.583  & 0.058 & 0.039 & 0.023 & 93.753 & 98.911 & 99.336 & 96.613 \\
Galician    & Indo-Euro  & 2.985  & 0.111 & 0.082 & 0.021 & 92.488 & 98.717 & 99.142 & 95.963 \\
Portuguese  & Indo-Euro  & 3.694  & 0.150 & 0.114 & 0.047 & 94.805 & 99.033 & 99.170 & 96.526 \\
French      & Indo-Euro  & 3.763  & 0.131 & 0.096 & 0.058 & 91.539 & 98.405 & 97.627 & 91.421 \\
Romanian    & Indo-Euro  & 6.043  & 0.172 & 0.118 & 0.056 & 90.492 & 98.291 & 98.957 & 94.913 \\
Turkish     & Turkic         & 6.536  & 0.122 & 0.071 & 0.057 & 93.939 & 98.831 & 99.284 & 96.338 \\
Lithuanian  & Indo-Euro  & 7.127  & 0.193 & 0.133 & 0.068 & 99.659 & 97.335 & 97.048 & 85.726 \\
Latin       & Indo-Euro  & 11.181 & 0.237 & 0.145 & 0.072 & 90.140 & 98.320 & 98.957 & 94.729 \\
Czech       & Indo-Euro  & 14.434 & 0.327 & 0.207 & 0.121 & 77.613 & 95.567 & 97.052 & 87.353 \\
Vietnamese  & Austroasiatic  & 25.234 & 0.783 & 0.586 & 0.513 & 67.787 & 92.369 & 91.475 & 74.152 \\
Bengali     & Indo-Euro  & 58.635 & 1.593 & 1.207 & 1.168 & --     & --     & 80.668 & 60.770 \\
Hebrew      & Afro-Asiatic   & 66.026 & 1.706 & 1.366 & 1.487 & --     & --     & 77.996 & 54.479 \\
Arabic      & Afro-Asiatic   & 82.017 & 1.430 & 1.350 & 1.784 & --     & --     & 67.064 & 35.637 \\
\bottomrule
\end{tabular}
\caption{Language-level averages of diacritic complexity metrics and diacritization performance, ordered by increasing diacritic density. 
For languages represented by multiple corpora, values are averaged across corpora. 
RS = Rune Surprisal; DTS = Diacritic Token Surprisal; DSS = Diacritic Structural Surprisal. 
Model performance is reported as word-level and rune-level accuracy for RNN and BERT-based diacritizers.}
\label{tab:avg_metrics_performance}
\end{table*}

\subsection{Correlation Between Diacritic Properties}

We first examine the inter-correlation between the proposed theoretical metrics.
For each corpus, we compute the mean token-level complexity score for each metric and present it in \autoref{tab:avg_metrics_performance}.

Across all corpora, the metrics are extremely highly correlated ($|r|>0.97$ for all pairwise comparisons), suggesting that at a broad cross-linguistic level they capture a shared underlying dimension of orthographic complexity.
However, when separating the languages by script type, clearer structural differences emerge.

In multi-diacritic scripts, Rune Surprisal (RS) and Diacritic Token Surprisal (DTS) remain strongly correlated ($r = 0.93$), since both are based directly on observed corpus frequencies.
Diacritic Structural Surprisal (DSS) is almost perfectly correlated with diacritic density ($r = 0.987$).
This means that in these systems, languages with more diacritics overall also tend to allow more distinct diacritic combinations.
Frequency-based measures correlate only moderately with structural measures ($r \sim 0.75$), suggesting that unpredictability of individual diacritics and the diversity of combination patterns are related but not identical properties.

In single-diacritic scripts, all metrics remain uniformly high ($r > 0.83$), with RS and DTS nearly identical ($r = 0.98$).
This reflects the fact that when only one diacritic may attach to a character, frequency-based and structural measures collapse into a single dominant dimension.

\subsection{Model Performance as a Function of Diacritic Properties}

We examine the relationship between corpus-level diacritic metrics and diacritization accuracy across all languages and corpora using the BERT-based model. \autoref{tab:correlations} reports Pearson correlations between each complexity metric and both word-level and rune-level accuracy.

Across all languages, word-level and rune-level accuracy for the BERT-based diacritizer shows a strong and highly significant negative correlation with diacritic complexity, which indicates that languages with higher orthographic complexity tend to score lower on restoration accuracy. This pattern holds across all metrics. Restricting the analysis to Indo-European languages show the same overall trend. This suggests that the effect is not limited to cross-family differences but also remains within a single language family.

\paragraph{Single-Diacritic Scripts.}

We next restrict the analysis to single-diacritic scripts, comprising 11 languages and 17 corpora. As observed above, in these scripts, the statistical metrics are mostly aligned.

For the BERT model, at both word and rune-level, DTS and DSS metrics show moderate correlations with accuracy. At rune-level, the Rune Suprisal metric is also moderately correlated. However, the overall effect is smaller and less significant than all languages.

In contrast, the RNN model exhibits strong and consistent negative correlations between diacritics complexity and accuracy across all metrics. At the word-level, correlations remain moderate, with RS showing the strongest association. This pattern suggests that the RNN architecture is more sensitive than BERT to increases in complexity even when structural variation is limited to a single diacritic per character.

\begin{table*}[t]
\centering
\small
\begin{tabular}{llrrrr}
\toprule
\textbf{Regime} & \textbf{Metric} & \textbf{BERT-Rune} & \textbf{BERT-Word} & \textbf{RNN-Rune} & \textbf{RNN-Word} \\
\midrule
\multicolumn{6}{l}{\textbf{Single-diacritic scripts}} \\
 & RS  & $-$0.47\phantom{***}  & \textbf{$-$0.55*\phantom{**}} & \textbf{$-$0.80***} & \textbf{$-$0.62**\phantom{*}} \\
 & DSS & \textbf{$-$0.50*\phantom{**}} & \textbf{$-$0.57*\phantom{**}} & \textbf{$-$0.76***} & \textbf{$-$0.57*\phantom{**}}  \\
\midrule
\multicolumn{6}{l}{\textbf{Multi-diacritic scripts}} \\
 & RS  & $-$0.59\phantom{***} & $-$0.50\phantom{***} & -- & -- \\
 & DSS & \textbf{$-$0.95**\phantom{*}} & \textbf{$-$0.92**\phantom{*}} & -- & -- \\
\midrule
\multicolumn{6}{l}{\textbf{Indo-European}} \\
 & RS  & \textbf{$-$0.98***} & \textbf{$-$0.97***} & -- & -- \\
 & DSS & \textbf{$-$0.99***} & \textbf{$-$0.96***} & -- & -- \\
\midrule
\multicolumn{6}{l}{\textbf{All Languages}} \\
 & RS  & \textbf{$-$0.94***} & \textbf{$-$0.94***} & -- & -- \\
 & DSS & \textbf{$-$0.99***} & \textbf{$-$0.98***} & -- & -- \\
\bottomrule
\end{tabular}
\caption{Pearson correlations between diacritization accuracy and diacritic complexity metrics: Rune Surprisal (RS) and Diacritic Structural Surprisal (DSS). Significance: * $p<0.05$, ** $p<0.01$, *** $p<0.001$.}
\label{tab:correlations}
\end{table*}

\paragraph{Multi-Diacritic Scripts.}

We next examine multi-diacritic scripts, comprising 4 languages and 6 corpora, numbers which affect the power of our analysis. In these systems, base characters may carry multiple simultaneous diacritics, and frequency-based and structural measures diverge.

For the BERT model, a clear pattern emerges at the word and rune level despite the small sample size. Structural measures show substantially stronger negative associations with accuracy than frequency-based measures. In contrast to the global and single-diacritic analyses, where the metrics largely behave similarly, structural complexity becomes the dominant predictor of performance. Although the analysis includes only six corpora, the direction of the effects is consistent with the findings so far.

This pattern suggests that, in multi-diacritic scripts, BERT performance is more strongly affected by structural complexity than by token-level unpredictability alone. In other words, the model is more sensitive to how many structurally distinct combinations there are than to how skewed the distribution of diacritics is.

Overall, the results indicate that multi-diacritic scripts introduce a qualitatively different form of orthographic complexity, which reveals that structural combinatorics plays a distinct and predictive role in model performance.

\subsection{Cross-Corpus Evaluation}

We examine cross-domain robustness within individual languages.
All non-Hebrew corpora were drawn from Universal Dependencies (UD; \citealp{de-marneffe-etal-2021-universal}), while the Hebrew corpora were taken from the fully diacritized dataset introduced by \citet{gershuni-pinter-2022-restoring}.

Cross-corpus evaluation was conducted for the following language pairs: Hebrew (literary / news), French (FTB, news / GSD, web), Spanish (AnCora, news / GSD, web), Turkish (Kenet, grammar examples / Penn, news), Romanian (RRT, news and web / SiMoNERo, medical), and Portuguese (CINTIL, mixed genres / GSD, web).
Across languages, cross-corpus evaluation shows a modest decrease in performance relative to in-domain results. On average, rune-level accuracy drops by approximately 1.5\%, and word-level accuracy by around 3\%. 
The drop is reasonable, as different domains can contain unique vocabularies and may vary in the distribution of character-diacritic combinations. These domain differences can affect restoration accuracy, particularly given the relatively limited size of each sampled corpus (approximately 300K characters) 
At the same time, the relatively small decrease suggests that diacritic usage is largely consistent within a language. Although specific words change across corpora, the patterns remain stable, allowing the model to generalize well across domains.

\section{Conclusion}

We presented the first comparative study of the orthographic phenomenon of diacritization, analyzed through statistical and information-theoretic constructs pertaining to the intrinsic properties of diacritized text across two dozen corpora.
We show that, as expected, the statistical presence and surprisal of diacritic marks across texts in a language affects the ability of state-of-the-art off-the-shelf diacritization models to restore marks in that language.
In future work, we will act on our actionable conclusions and provide practical recommendations for diacritics restoration and for handling undiacritized text in downstream NLP systems, building also on the principles demarcated by \citet{gorman-pinter-2025-dont}.

An additional avenue for future work would be to enrich the presented study by incorporating additional features for correlation analysis, such as the functional role of diacritics in a language.
For example, diacritics may encode phonological, morphological, or tonal information; these differences can help explain variation in model performance beyond structural complexity alone.

\section*{Acknowledgements}

We would like to thank Kyle Gorman and the anonymous reviewers for valuable comments on earlier drafts.
This research was supported by grant no.~2022215 from the United States--Israel Binational Science Foundation (BSF), Jerusalem, Israel.

\section{Bibliographical References}\label{sec:reference}

\bibliographystyle{lrec2026-natbib}
\bibliography{anthology-1,anthology-2,lrec2026-example}

\begin{thebibliography}{14}
\expandafter\ifx\csname natexlab\endcsname\relax\def\natexlab#1{#1}\fi

\bibitem[{Belinkov and Glass(2015)}]{belinkov-glass-2015-arabic}
Yonatan Belinkov and James Glass. 2015.
\newblock \href {https://doi.org/10.18653/v1/D15-1274} {{A}rabic diacritization with recurrent neural networks}.
\newblock In \emph{Proceedings of the 2015 Conference on Empirical Methods in Natural Language Processing}, pages 2281--2285, Lisbon, Portugal. Association for Computational Linguistics.

\bibitem[{Clark et~al.(2022)Clark, Garrette, Turc, and Wieting}]{clark-etal-2022-canine}
Jonathan~H. Clark, Dan Garrette, Iulia Turc, and John Wieting. 2022.
\newblock \href {https://doi.org/10.1162/tacl_a_00448} {Canine: Pre-training an efficient tokenization-free encoder for language representation}.
\newblock \emph{Transactions of the Association for Computational Linguistics}, 10:73--91.

\bibitem[{Darwish et~al.(2017)Darwish, Mubarak, and Abdelali}]{darwish-etal-2017-arabic}
Kareem Darwish, Hamdy Mubarak, and Ahmed Abdelali. 2017.
\newblock \href {https://doi.org/10.18653/v1/W17-1302} {{A}rabic diacritization: Stats, rules, and hacks}.
\newblock In \emph{Proceedings of the Third {A}rabic Natural Language Processing Workshop}, pages 9--17, Valencia, Spain. Association for Computational Linguistics.

\bibitem[{de~Marneffe et~al.(2021)de~Marneffe, Manning, Nivre, and Zeman}]{de-marneffe-etal-2021-universal}
Marie-Catherine de~Marneffe, Christopher~D. Manning, Joakim Nivre, and Daniel Zeman. 2021.
\newblock \href {https://doi.org/10.1162/coli_a_00402} {{U}niversal {D}ependencies}.
\newblock \emph{Computational Linguistics}, 47(2):255--308.

\bibitem[{Devlin et~al.(2019)Devlin, Chang, Lee, and Toutanova}]{devlin-etal-2019-bert}
Jacob Devlin, Ming-Wei Chang, Kenton Lee, and Kristina Toutanova. 2019.
\newblock \href {https://doi.org/10.18653/v1/N19-1423} {{BERT}: Pre-training of deep bidirectional transformers for language understanding}.
\newblock In \emph{Proceedings of the 2019 Conference of the North {A}merican Chapter of the Association for Computational Linguistics: Human Language Technologies, Volume 1 (Long and Short Papers)}, pages 4171--4186, Minneapolis, Minnesota. Association for Computational Linguistics.

\bibitem[{Gershuni and Pinter(2022)}]{gershuni-pinter-2022-restoring}
Elazar Gershuni and Yuval Pinter. 2022.
\newblock \href {https://doi.org/10.18653/v1/2022.findings-naacl.75} {Restoring {H}ebrew diacritics without a dictionary}.
\newblock In \emph{Findings of the Association for Computational Linguistics: NAACL 2022}, pages 1010--1018, Seattle, United States. Association for Computational Linguistics.

\bibitem[{Goldhahn et~al.(2012)Goldhahn, Eckart, and Quasthoff}]{goldhahn-etal-2012-building}
Dirk Goldhahn, Thomas Eckart, and Uwe Quasthoff. 2012.
\newblock \href {https://aclanthology.org/L12-1154/} {Building large monolingual dictionaries at the {L}eipzig corpora collection: From 100 to 200 languages}.
\newblock In \emph{Proceedings of the Eighth International Conference on Language Resources and Evaluation ({LREC}'12)}, pages 759--765, Istanbul, Turkey. European Language Resources Association (ELRA).

\bibitem[{Gorman and Pinter(2025)}]{gorman-pinter-2025-dont}
Kyle Gorman and Yuval Pinter. 2025.
\newblock \href {https://doi.org/10.18653/v1/2025.naacl-short.25} {Don{'}t touch my diacritics}.
\newblock In \emph{Proceedings of the 2025 Conference of the Nations of the Americas Chapter of the Association for Computational Linguistics: Human Language Technologies (Volume 2: Short Papers)}, pages 285--291, Albuquerque, New Mexico. Association for Computational Linguistics.

\bibitem[{Krstev et~al.(2018)Krstev, Stankovi{\'c}, and Vitas}]{krstev-etal-2018-knowledge}
Cvetana Krstev, Ranka Stankovi{\'c}, and Du{\v{s}}ko Vitas. 2018.
\newblock \href {https://aclanthology.org/2018.clib-1.7/} {Knowledge and rule-based diacritic restoration in {S}erbian}.
\newblock In \emph{Proceedings of the Third International Conference on Computational Linguistics in Bulgaria (CLIB 2018)}, pages 41--51, Sofia, Bulgaria. Department of Computational Linguistics, Institute for Bulgarian Language, Bulgarian Academy of Sciences.

\bibitem[{N\'{a}plava et~al.(2021)N\'{a}plava, Straka, and Strakov\'{a}}]{naplava-straka-strakova-2021}
Jakub N\'{a}plava, Milan Straka, and Jana Strakov\'{a}. 2021.
\newblock \href {https://doi.org/10.14712/00326585.013} {{Diacritics Restoration using BERT with Analysis on Czech language}}.
\newblock \emph{The Prague Bulletin of Mathematical Linguistics}, 116:27--42.

\bibitem[{N{\'a}plava et~al.(2018)N{\'a}plava, Straka, Stra{\v{n}}{\'a}k, and Haji{\v{c}}}]{naplava-etal-2018-diacritics}
Jakub N{\'a}plava, Milan Straka, Pavel Stra{\v{n}}{\'a}k, and Jan Haji{\v{c}}. 2018.
\newblock \href {https://aclanthology.org/L18-1247/} {Diacritics restoration using neural networks}.
\newblock In \emph{Proceedings of the Eleventh International Conference on Language Resources and Evaluation ({LREC} 2018)}, Miyazaki, Japan. European Language Resources Association (ELRA).

\bibitem[{Shmidman et~al.(2020)Shmidman, Shmidman, Koppel, and Goldberg}]{shmidman-etal-2020-nakdan}
Avi Shmidman, Shaltiel Shmidman, Moshe Koppel, and Yoav Goldberg. 2020.
\newblock \href {https://doi.org/10.18653/v1/2020.acl-demos.23} {{N}akdan: Professional {H}ebrew diacritizer}.
\newblock In \emph{Proceedings of the 58th Annual Meeting of the Association for Computational Linguistics: System Demonstrations}, pages 197--203, Online. Association for Computational Linguistics.

\bibitem[{Tsarfaty et~al.(2019)Tsarfaty, Sadde, Klein, and Seker}]{tsarfaty-etal-2019-whats}
Reut Tsarfaty, Shoval Sadde, Stav Klein, and Amit Seker. 2019.
\newblock \href {https://doi.org/10.18653/v1/D19-3044} {What{'}s wrong with {H}ebrew {NLP}? and how to make it right}.
\newblock In \emph{Proceedings of the 2019 Conference on Empirical Methods in Natural Language Processing and the 9th International Joint Conference on Natural Language Processing (EMNLP-IJCNLP): System Demonstrations}, pages 259--264, Hong Kong, China. Association for Computational Linguistics.

\bibitem[{Zerrouki and Balla(2017)}]{zerrouki2017tashkeela}
Taha Zerrouki and Amar Balla. 2017.
\newblock Tashkeela: Novel corpus of arabic vocalized texts, data for auto-diacritization systems.
\newblock \emph{Data in Brief}, 11:147--151.

\end{thebibliography}


\end{document}